\documentclass[conference,10pt]{IEEEtran}
\usepackage{epsfig,rotating,setspace,latexsym,amsmath,epsf,amssymb,amsfonts,bm,theorem,subfigure,epstopdf}
\usepackage{cite,authblk}
\usepackage{bbm}
\usepackage{color}
\usepackage{mathtools}
\usepackage{algorithm}
\usepackage{algpseudocode}
\usepackage{verbatim}
\usepackage{xcolor}

\algrenewcommand\algorithmicforall{\textbf{foreach}}
\algrenewcommand\algorithmicindent{.8em}

\algnewcommand\algorithmicforeach{\textbf{for each}}
\algdef{S}[FOR]{ForEach}[1]{\algorithmicforeach\ #1\ \algorithmicdo}

\IEEEoverridecommandlockouts
\allowdisplaybreaks

\begin{document}

\title{Re-ranking the Context for Multimodal Retrieval Augmented Generation}

\author{
Matin Mortaheb$^{\dag}$, Mohammad A. (Amir) Khojastepour$^{*}$, Srimat T. Chakradhar$^{*}$, Sennur Ulukus$^{\dag}$ \\
\normalsize $^{\dag}$University of Maryland, College Park, MD, $^{*}$NEC Laboratories America, Princeton, NJ \\
\normalsize \emph{mortaheb@umd.edu, amir@nec-labs.com, chak@nec-labs.com, ulukus@umd.edu}
}

% \author{
% Matin Mortaheb\\
% \normalsize University of Maryland, College Park\\
% \normalsize College Park, MD\\
% \normalsize \emph{mortaheb@umd.edu}
% \and
% Mohammad A. (Amir) Khojastepour\\
% \normalsize NEC Laboratories, America\\
% \normalsize Princeton, NJ\\
% \normalsize \emph{amir@nec-labs.com}
% \and
% Srimat T. Chakradhar\\
% \normalsize NEC Laboratories, America\\
% \normalsize Princeton, NJ\\
% \normalsize \emph{chak@nec-labs.com}
% \and
% Sennur Ulukus\\
% \normalsize University of Maryland, College Park\\
% \normalsize College Park, MD\\
% \normalsize \emph{ulukus@umd.edu}
% }
\maketitle

\begin{abstract}
Retrieval-augmented generation (RAG) enhances large language models (LLMs) by incorporating external knowledge to generate a response within a context with improved accuracy and reduced hallucinations. However, multi-modal RAG systems face unique challenges: (i) the retrieval process may select irrelevant entries to user query (e.g., images, documents), and (ii) vision-language models or multi-modal language models like GPT-4o may hallucinate when processing these entries to generate RAG output. In this paper, we aim to address the first challenge, i.e, improving the selection of relevant context from the knowledge-base in retrieval phase of the multi-modal RAG. Specifically, we leverage the relevancy score (RS) measure designed in our previous work for evaluating the RAG performance to select more relevant entries in retrieval process. The retrieval based on embeddings, say CLIP-based embedding, and cosine similarity usually perform poorly particularly for multi-modal data. We show that by using a more advanced relevancy measure, one can enhance the retrieval process by selecting more relevant pieces from the knowledge-base and eliminate the irrelevant pieces from the context by adaptively selecting up-to-$k$ entries instead of fixed number of entries. Our evaluation using COCO dataset demonstrates significant enhancement in selecting relevant context and accuracy of the generated response. 
% Our evaluation, conducted on a dataset derived from the COCO benchmark, demonstrates that this approach significantly enhances the relevance of the retrieval compared to CLIP, resulting in more accurate and contextually appropriate responses, as measured by our designed correctness score (CS). These findings highlight the effectiveness of the RS model in improving both the selection and generation phases of multimodal RAG systems, offering a robust framework for mitigating hallucinations and enhancing response quality.
\end{abstract}


\section{Introduction}
\label{sec:intro}

Retrieval-augmented generation (RAG) \cite{lewis2020retrieval} systems enhance large language models (LLMs) \cite{achiam2023gpt} by incorporating external knowledge to generate coherent responds based on a given context, improve the response accuracy and reduce hallucinations. However, the quality of the generated response in RAG systems heavily depends on the retrieval process. Selecting the most relevant data from a database based on the user query is essential for the system to generate accurate and contextually appropriate response. A common approach for retrieval in RAG is top-$k$ selection by first ranking the entries from the knowledge-base based on similarity scores between their embeddings and the user query and then selecting the top-$k$ entries. For image and text retrieval, CLIP-based methods \cite{radford2021learning} are widely used due to their strong performance in aligning image and text embeddings across modalities. While effective in capturing general semantic similarity, CLIP struggles to detect irrelevancy as accurately as relevancy, making it less reliable for context-sensitive tasks.

\begin{figure}[t]
\centerline{\includegraphics[width=1\linewidth]{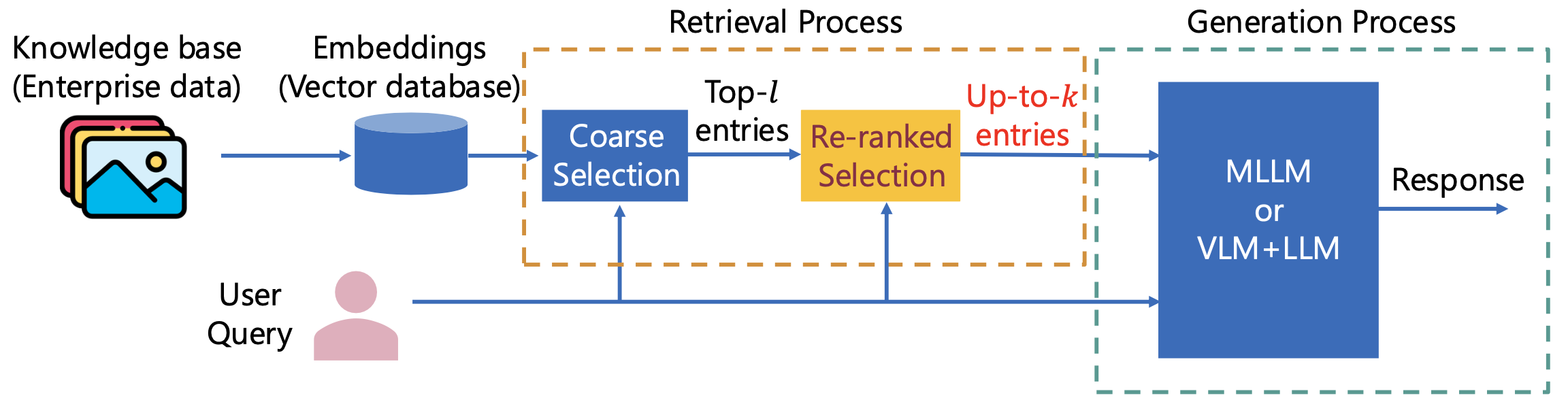}}
  \caption{Re-ranking via relevancy score (RS).}
\centering
\label{fig:reranking_framework}
\vspace*{-0.5cm}
\end{figure}

One fundamental limitation of CLIP is its tendency to assign high similarity scores to visually or semantically generic content, even when it is irrelevant to the query. For example, consider a query like ``a child playing soccer". A CLIP-based retrieval system might assign high similarity to images of any child or soccer scene, even if the images lack both elements together. This inability to effectively filter out irrelevant data can lead to suboptimal retrieval, and consequently, hallucinations in downstream tasks. In multi-modal RAG systems, where accuracy depends on precise alignment between the retrieved data and the user query, such limitations can significantly degrade the system's reliability.

Re-ranking techniques offer a promising solution to refine initial retrieval results by re-evaluating the relevance of retrieved entries before passing them to the response generation phase. Recent works in multimodal retrieval have introduced advanced re-ranking methods to enhance relevance estimation \cite{chen2024mllm, qu2023learnable, yang2024toward, xu2023ai, messina2022aladin}. For example, the work in \cite{qu2023learnable} reconstructs data samples based on top-ranked intra- and inter-modal neighbors (referred to as ``pillars'') to improve retrieval accuracy. The work in \cite{chen2024mllm} employs multi-modal large language models (MLLMs) with knowledge-enhanced re-ranking and noise-injected training to refine retrieval results. Also, the work in \cite{yang2024toward} demonstrates the potential of vision-language models for relevance evaluation, but it highlights these models’ tendency to over-rely on semantic similarity, often failing in cases that require precise contextual understanding. A common limitation of most re-ranking approaches is their reduced ability to detect irrelevant information, as MLLMs are primarily trained on datasets containing relevant image-query pairs. Moreover, using VLM or MLLM in prior art for re-ranking is limited to a binary decision, i.e., ``YES/NO''. In contrast, our RS model addresses these challenges by being explicitly trained to assess context-specific relevance, effectively filtering out irrelevant content and ensuring that retrieved entries align with the user’s intent. Moreover, analogous to similarity score, the RS model provides a quantitative value that can be used to explicitly rank entries based on their relevancy as well as eliminating the entries due to irrelevancy.  

In this work, we address the shortcomings of CLIP-based retrieval by proposing the use of previously developed RS model for re-ranking retrieved entries in multi-modal RAG systems \cite{mortahebRAGCheck}. Unlike CLIP, the RS model is specifically trained to assess query-specific relevance while penalizing irrelevant entries effectively. For example, given the query ``a doctor holding a medical instrument", RS can prioritize images depicting this exact context over generic depictions of doctors or medical tools, which CLIP might incorrectly rank highly. By replacing the initial CLIP-based top-$k$ selection process with RS-based re-ranking, our method ensures that retrieved data aligns more closely with the user’s intent. Our evaluation, conducted on a dataset derived from the COCO benchmark, demonstrates that RS-based re-ranking significantly improves the quality of selected images, leading to more accurate and factually grounded responses as measured by our correctness score (CS). Furthermore, we highlight the broader implications of using advanced relevance metrics in retrieval refinement for multi-modal RAG systems, showcasing their potential to mitigate hallucinations and enhance response reliability. The design, training, and evaluation of both RS and CS models are available in \cite{mortahebRAGCheck}.

\section{Multi-modal RAG}
In RAG, the knowledge-base is pre-processed by generating embeddings for each piece of information which is stored in a vector database enabling fast similarity-based retrieval. In multi-modal RAG systems, embeddings for different data types, such as text and images, are derived using modality-specific encoders which share the same embedding space that is used for embedding the query. For instance, CLIP embeddings can be used, where text and image data are encoded using the CLIP text and vision encoders, respectively, and relevance is determined through cosine similarity. It is noted that even though by using CLIP for both image and text, the embedding spaces are the same, the comparison is not seamless, i.e., the range of similarity between text-text pair is quite different with that of text-image pair. This is yet another challenge in selecting the relevant pieces of information for the scenarios where both image and texts are selected from the vector database.

The retrieved top-$k$ entries in the first phase of the RAG, forming the \emph{raw context}, are passed to the next phase of the RAG, i.e., the generation module, to produce the final response. In multi-modal scenarios, RAG systems may utilize different approaches for the generation module. One approach involves using vision language models (VLMs) \cite{liu2024visual, lin2024vila} to convert the retrieved image context into text-based descriptions, which are then combined with the query to generate the final response. Alternatively, MLLMs can directly process the retrieved images along with the user query to generate the response.

\section{RS score vs CLIP for RAG selection}
The RS (please see details in \cite{mortahebRAGCheck}) is a specialized metric designed to evaluate the relevancy of a piece of information from the vector database to the user query in multi-modal RAG systems. The RS model is designed by using a VLM with specific fine-tuned head that learns the semantic relationship between a user query and a retrieved entry, such as an image or a text document. The training of the RS model leverages a carefully curated dataset that includes a human-annotated dataset as well as synthetic query-context pairs generated by ChatGPT partially verified by human evaluators. We build a balanced dataset where each entry is a triplet comprising an image, and a pair of positive and negative statements with respect to the image. The dataset is partitioned into a training dataset of 121,000 triplets and an evaluation dataset of 2000 triplets. The training on a balanced dataset ensures that the RS model captures not only the general semantic alignment but also fine-grained contextual relevance/irrelevance.

To compute RS, the RS model takes the query and an entry (e.g., an image or text) as an input and produce a scalar score between 0 and 1 as an output. The higher the score, the higher the relevancy, i.e., the score 1 is the highest relevancy and 0 is the ultimate irrelevancy. By design, the RS score is naturally normalized by using a sigmoid activation function at the last layer of our fine-tuned head. % which facilitate comparison across entries. 
During training, the RS model minimizes a modified version of RLHF loss function that penalizes mismatched query-context pairs while rewarding alignment with the most relevant entries. This enables the RS model to differentiate between truly relevant data and entries that might exhibit superficial similarity to the query. 

% During the inference, given a pair of $(\mathcal{I}, q)$ where $\mathcal{I}$ is an image and $q$ is a text-based query, we can obtain the RS as:  
% \begin{equation}
%     \text{RS} = \sigma\left(y_{-1}(\mathcal{I},q)\right)
% \end{equation}

%Fig.~\ref{fig:rs_histogram} illustrates that the RS evaluated over an evaluation set of image-statement pairs produce the full range of $(0,1)$ scores while CLIP has limited range as seen in Fig.\ref{fig:clip_histogram}.

Compared to CLIP-based methods, which rely on cosine similarity between embeddings to rank image-text pairs, RS is designed to excel in detecting both relevance and irrelevance. While CLIP is effective at capturing general semantic similarity, it often assigns high scores to visually or conceptually generic data, even if it is not contextually relevant to the query. For instance, given a query like “a person reading a book in a park”, CLIP may retrieve images of people in a park or people reading indoors, failing to prioritize those that align with the full context of the query. In contrast, RS is specifically tuned to penalize such irrelevant data, enabling more precise selection of relevant entries.

\begin{figure}[t]
\centerline{\includegraphics[width=0.9\linewidth]{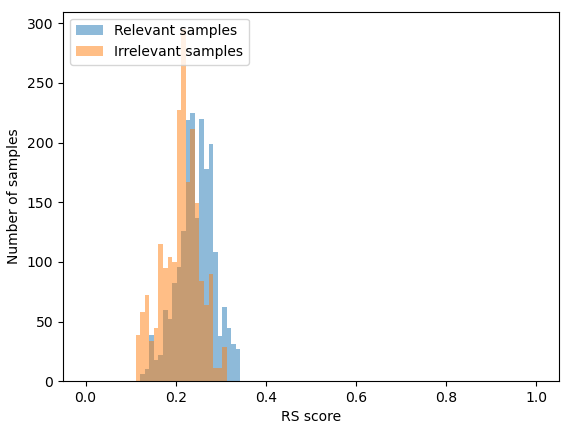}}
  \caption{CLIP-score histogram.}
\centering
\label{fig:clip_histogram}
\vspace*{-0.3cm}
\end{figure}

We evaluate both CLIP and our RS model by using 2000 pairs of images and positive statements as well as 2000 pairs of the same images and negative statements drawn from evaluation dataset. Fig.~\ref{fig:clip_histogram} shows the histograms of similarity scores based on CLIP for 2000 evaluation dataset. The histogram for the similarity scores between the CLIP embeddings of the positive (negative) statements and the image are depicted by blue (orange) color and labeled as relevant (irrelevant) statements. Let us define \emph{CLIP-score} as the similarity score based on CLIP embeddings. From the histogram, it is evident that (i) the CLIP-score between image-text pairs has limited range, e.g., about $(0.13,0.35)$, and (ii) the distribution of the CLIP-score for relevant and irrelevant statements has considerable overlap which means that such similarity score struggles to adequately distinguish between relevant and irrelevant pieces of data. 

\begin{figure}[t]
\centerline{\includegraphics[width=0.9\linewidth]{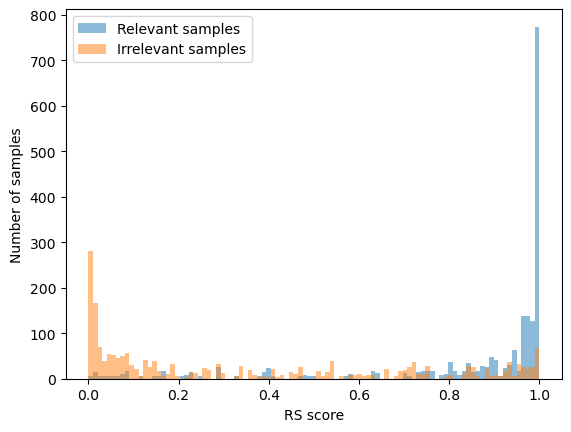}}
  \caption{RS score histogram.}
\centering
\label{fig:rs_histogram}
\vspace*{-0.3cm}
\end{figure}

In contrast, as shown in Fig.~\ref{fig:rs_histogram}, the RS score on the same evaluation dataset effectively shifts the distribution, creating a clearer separation between relevant and irrelevant samples. In particular, the RS score (i) exploits the full output range of $(0,1)$ and (ii) in comparison to the similarity score based on the CLIP, the two distributions for relevant and irrelevant samples based on RS are well separated. As a simple measure of separation one can consider the distance between the mean of relevant and irrelevant samples for CLIP-score and RS that are calculated as 0.03 and 0.41 for the distribution given in Fig.~\ref{fig:clip_histogram} and Fig.~\ref{fig:rs_histogram}, respectively, while the maximum separation between the mean of such scores that are limited in interval [0,1] is at most 1. A more suitable measure of separation between two discrete distribution is Jensen-Shannon divergence (JSD) which is calculated as 0.32 and 0.58 for CLIP-score and RS, respectively, with the upper bound 0.83 for scores that are limited in interval [0,1].

\begin{figure}[t]
\centerline{\includegraphics[width=0.9\linewidth]{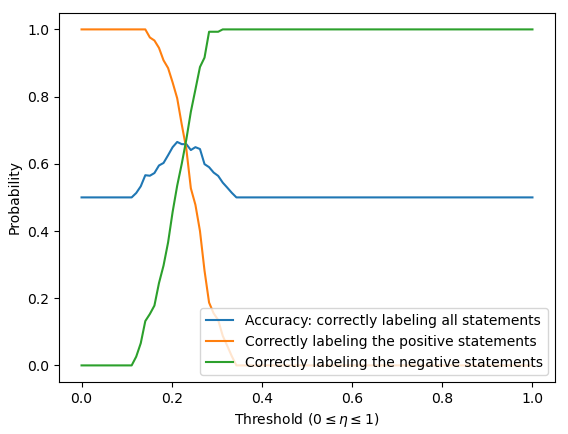}}
  \caption{CDF for relevant (irrelevant) samples in CS-score.}
\centering
\label{fig:clip_threshold}
\vspace*{-0.3cm}
\end{figure}

The threshold may be used on the value obtained by CLIP-score in order to make a binary classification of the statements into relevant/irrelevant. In order to easily interpret the performance of CLIP-score based on the different thresholds we plot the cumulative density function (cdf) of the distribution of the irrelevant samples depicted in Fig.~\ref{fig:clip_histogram} and 1-cdf of the distribution of the relevant samples as illustrated in Fig.~\ref{fig:clip_threshold}. By selecting a threshold value on the CLIP-score in the horizontal axis, one can easily see the probability of correct labeling of positive and negative statements by reading the corresponding values of the plotted curves. The overall accuracy is also the average between these two curves denoted by the legend `accuracy'. It is observed that the maximum accuracy for the CLIP-score on the evaluated dataset is about 0.65 occurring at the threshold of about 0.2.

\begin{figure}[t]
\centerline{\includegraphics[width=0.9\linewidth]{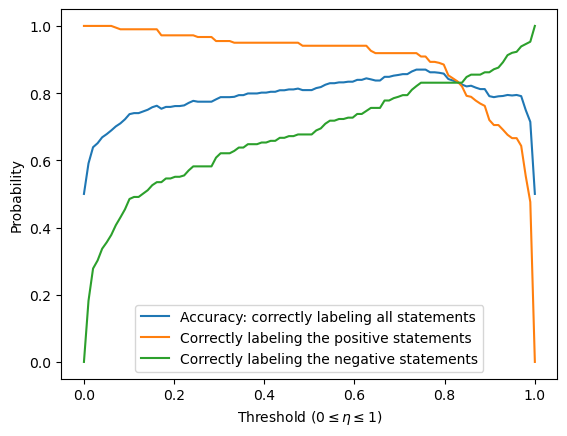}}
  \caption{CDF for relevant (irrelevant) samples in RS score.}
\centering
\label{fig:rs_threshold}
\vspace*{-0.3cm}
\end{figure}

Similarly, we plot the cdf and 1-cdf of the distributions of the irrelevant and relevant samples, respectively, for the RS in Fig.~\ref{fig:rs_threshold}. By selecting an optimized threshold, RS achieves significantly higher accuracy of about 0.88 occurring at threshold 0.75. 

%This figure demonstrates that over a wide range of selected thresholds, the probabilities of true detection for positive (relevant) and negative (irrelevant) samples remain nearly identical, highlighting CLIP's inability to reliably separate the two classes.

% Moreover, as depicted in Fig.~\ref{fig:rs_threshold}, by selecting an optimized threshold, RS achieves significantly higher confidence in selecting positive (relevant) samples compared to CLIP.

\section{Re-ranking with RS Score}

In multi-modal RAG systems, the retrieval process plays a critical role in ensuring that selected data aligns with the user query. While CLIP-based methods are efficient for initial retrieval, they often fall short in distinguishing between truly relevant and irrelevant entries. One may directly use RS score to enhance retrieval process. However, this would elongate the retrieval by a factor of $\times35$. To address this, we propose the following re-ranking mechanism.

For selecting up-to-$k$ entries, we first retrieve a larger candidate set of size $l$, $l > k$, using the computationally efficient CLIP-score. These $l$ candidates are then re-ranked based on their RS scores, which incorporate the relevance of each candidate to the query more precisely. Finally, the up-to-$k$ entries from the re-ranked list are selected as the final retrieved entries.

Our proposed up-to-$k$ algorithm works as follows. From the list of $l$ candidates, we eliminate the entries with RS below threshold $\tau_{lo}$ and keep the entries with RS above $\tau_{hi}$. Let $\tau$ be the median of the RS for the rest of the entries. We discard the entries below $\tau$ and keep the entries above $\tau$. If the number of kept entries exceeds $k$, we select $k$ entries with highest RS, otherwise we select only the kept entries. 

%To further improve the quality of the selection, we introduce an adaptive mechanism that selects up to $k$ candidates based on a threshold $\tau$. Instead of always selecting exactly $k$ images, this approach ensures that only candidates with RS scores above $\tau$ are included. For example, consider a query such as "a surgeon performing heart surgery." If only one image in the retrieved set is highly relevant, selecting $k = 5$ images would force the inclusion of four irrelevant images, potentially introducing hallucinations into the generated response. By applying the threshold $\tau$, the system dynamically adjusts the number of selected entries, including only those that meet the relevance criterion.

This adaptive re-ranking method not only improves the alignment of retrieved entries with the user query but also reduces the likelihood of irrelevant data that could cause hallucinations in downstream tasks. By combining the time efficiency of CLIP for fast initial retrieval with enhanced capability of RS in measuring the relevancy, our approach proposes a balance between computational complexity and retrieval quality. Our experimental results illustrates significant enhancement not only in retrieval but also in the accuracy of the final output generated by multi-modal RAG systems.

\section{Numerical Results}\label{sec:numerical-results}

We use RS and CS model developed in \cite{mortahebRAGCheck} for evaluation. The performance of RS and CS has shown 88\% and 91\% alignment with human evaluation, respectively, as reported in \cite{mortahebRAGCheck} over 5000 human evaluated samples of RAG queries/responses built on a set of 1281 images from COCO dataset. Here, our evaluation relies on the reported accuracy of the corresponding RS and CS models.

%Here, we first compare the retrieval relevancy performance of top-$k$ selection for $k=5$ using CLIP-score for various CLIP models and up-to-$k$ selection directly using RS or re-ranking with $l=10, 20$.

\begin{figure}[t]
\centerline{\includegraphics[width=0.9\linewidth]{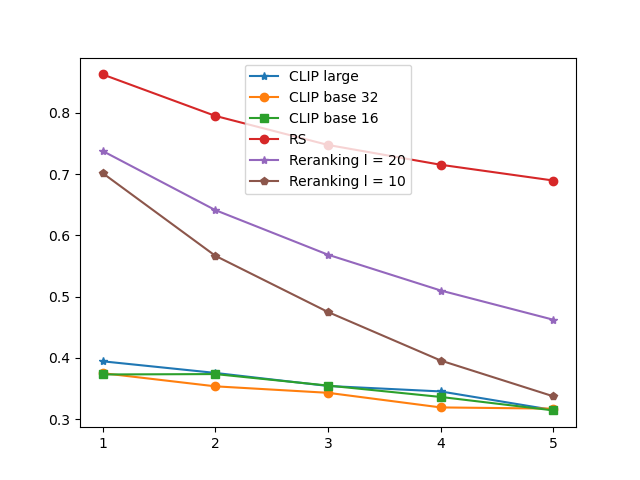}}
  \caption{RAG retrieval relevancy performance.}
\centering
\label{fig:rs_comparison}
\vspace*{-0.3cm}
\end{figure}

Here, we first compare the retrieval relevancy performance of CLIP-score with directly using RS and re-ranking with $l=10, 20$. We randomly select 1,000 questions from the COCO-QA dataset and use the same set of 1,281 images from the COCO dataset. Fig.~\ref{fig:rs_comparison} illustrates the average relevancy score (RS) for the top 5 selected entries where the horizontal axis is the order of the selected entry for each method. We observe that different CLIP models (large, base-16, and base-32) perform similarly, while using RS directly, the relevancy is boosted almost by a factor of 2. However, this boost comes at a cost of 35 times slower retrieval process. The re-ranking mechanism with $l=20$ and $l=10$ provide a significant boost in relevancy performance of about 85\% and 71\%, at cost of 1.55 and 1.27 times slower retrieval process, respectively.

%In this section, we evaluate our proposed re-ranking technique with RS with CLIP-score dot product method in terms of how well it results in selecting the better context and generating more correct response in different RAG schemes. 

%For calculating the RS score, we use the RS model introduced in \cite{mortahebRAGCheck}. First, based on the evaluation in \cite{mortahebRAGCheck}, the RS and CS scores have 88-91\% alignment with human evaluation. Therefore, RS and CS scores are good candidate to measure the the relevance in terms of RAG selection and correctness in terms of RAG generation. 

%We randomly select 1,000 questions from the COCO-QA dataset and gather 1,281 random images from the COCO dataset. For each question, we select the 5 most relevant images from the dataset using different methods. The first method is the conventional approach, which uses cosine similarity with the CLIP model. The second method calculates RS values for all question-image pairs.

%As shown in Fig.~\ref{fig:rs_comparison}, using RS values alone for retrieving data results in selecting nearly 50\% more relevant images. However, this process is approximately 35 times slower. To address this inefficiency, we adopt the proposed reranking method using RS values. First, we obtain $l=10, 20$ candidate images from the CLIP method. Then, we calculate the RS values among these candidates and select the top 5 images. On average, the 5 selected images exhibit at least 30\% better relevance to the user query.

\begin{figure}[t]
\centerline{\includegraphics[width=0.9\linewidth]{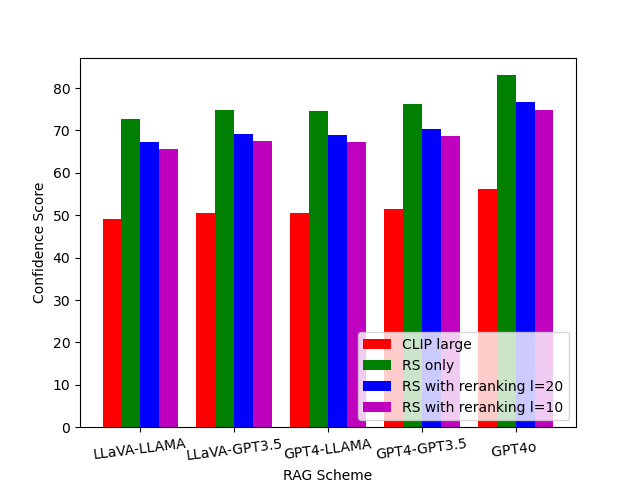}}
  \caption{RAG generated output performance.}
\centering
\label{fig:cs_comparison}
\vspace*{-0.3cm}
\end{figure}

Next, we compare the generated output performance of different RAG implementations, i.e., four RAG schemes with combinations of different VLMs and LLMs as well as a RAG scheme directly based on GPT-4o that processes multi-modal context, e.g., multiple images, and text query simultaneously. For each RAG implementation, we use 4 retrieval methods: top-$k$ selection for $k=5$ using CLIP-large, up-to-$k$ selection directly using RS, and up-to-$k$ selection using re-ranking with $l=10, 20$. We use $\tau_{lo} = 0.3$ and $\tau_{hi} = 0.75$ for the up-to-$k$ algorithm. The performance of the generated output is measured in terms of confidence score that is calculated as the geometric mean of CS and RS. The confidence score incorporates both the relevancy of the retrieved context and accuracy of the output in the view of the context in a single measure. It is important to note that a RAG may generate an output that is correct within the selected context, i.e., CS is high, but the retrieved context is not relevant, i.e., RS is low, hence, the output may not contain the answer to the user query. Fig.~\ref{fig:cs_comparison} indicates that applying RS directly or through re-ranking improves response quality across all evaluated RAG structures.

%Fig.~\ref{fig:cs_comparison} shows the confidence score when different selection methods are applied to various RAG structures. We evaluated 5 RAG structures. In 4 of these structures, a Vision-Language Model (VLM) is used to convert the retrieved images into text-based context, and a Large Language Model (LLM) combines the information and generates the response. In the fifth structure, GPT directly processes all the retrieved documents along with the query to generate the final response. 

%The confidence score is calculated as the geometric mean of the correctness and relevance scores. We focus on the confidence score because a final answer may sometimes be entirely correct but irrelevant to the user query. By incorporating both RS and CS, the confidence score reflects how well the retrieved context aligns with the user query and how accurate the final answer is concerning the retrieved context. 

%Our results indicate that applying RS scores improves response quality across all evaluated RAG structures. Additionally, we implemented the up-to-$k$ technique in reranking to assess its impact on the confidence score. The results demonstrate that selecting only a small portion of the most relevant data leads to improved answers.

\section{Conclusion}
We presented a retrieval process for RAG comprising re-ranking and up-to-$k$ selection method. Our proposed solution (i) reduces the hallucinations by selecting more relevant entries from the knowledge-base, and (ii) improves the relevancy of the output to the query. 

\bibliographystyle{unsrt}
\bibliography{reference}
\end{document}